\documentclass{article}

\usepackage{arxiv}
\usepackage{datetime}
\usepackage[utf8]{inputenc} 
\usepackage[T1]{fontenc}    
\usepackage{hyperref}       
\usepackage{url}            
\usepackage{booktabs}       
\usepackage{amsfonts}       
\usepackage{nicefrac}       
\usepackage{microtype}      
\usepackage{lipsum}
\usepackage{amsmath}
\usepackage{multirow}
\usepackage{graphicx}
\graphicspath{{Images/}}

\title{\textbf{A Deep Convolutional Neural Networks Based Multi-Task Ensemble Model for Aspect and Polarity Classification in Persian Reviews  }}

\author{ \href{https://orcid.org/0000-0002-4008-7820}{\includegraphics[scale=0.06]{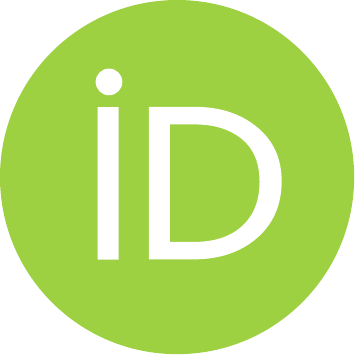}\hspace{1mm}Milad Vazan}\thanks{https://github.com/miladvazan/Convolutional-Neural-Networks-Based-Multi-task-Ensemble-Model } \\
		\\Department of Computer Science\\
\\ Faculty of Mathematics, Statistics, and Computer Science, University of Tabriz, Tabriz, Iran\\
 vazanmilad@gmail.com \\
	\And
	\href{https://orcid.org/0000-0000-0000-0000}{\includegraphics[scale=0.06]{orcid.pdf}\hspace{1mm}Fatemeh Sadat Masoumi} \
	\\Department of Computer Science\\
\\Faculty of Mathematics, Statistics, and Computer Science, University of Allameh Tabataba'i, Tehran, Iran\\
	\And
	\href{https://orcid.org/0000-0000-0000-0000}{\includegraphics[scale=0.06]{orcid.pdf}\hspace{1mm}Sepideh Saeedi Majd} \\
	\\Department of Computer Science\\
\\ Faculty of Mathematics, Statistics, and Computer Science, University of Tabriz, Tabriz, Iran\\
}

\begin{document}
\maketitle
\begin{abstract} \normalsize

Aspect-based sentiment analysis is of great importance and application because of its ability to identify all aspects discussed in the text. However, aspect-based sentiment analysis will be most effective when, in addition to identifying all the aspects discussed in the text, it can also identify their polarity. Most previous methods use the pipeline approach, that is, they first identify the aspects and then identify the polarities. Such methods are unsuitable for practical applications since they can lead to model errors. Therefore, in this study, we propose a multi-task learning model based on Convolutional Neural Networks (CNNs), which can simultaneously detect aspect category and detect aspect category polarity. creating a model alone may not provide the best predictions and lead to errors such as bias and high variance. To reduce these errors and improve the efficiency of model predictions, combining several models known as ensemble learning may provide better results. Therefore, the main purpose of this article is to create a model based on an ensemble of multi-task deep convolutional neural networks to enhance sentiment analysis in Persian reviews. We evaluated the proposed method using a Persian language dataset in the movie domain. Jacquard index and Hamming loss measures were used to evaluate the performance of the developed models. The results indicate that this new approach increases the efficiency of the sentiment analysis model in the Persian language.

\end{abstract}


\vfill
\section{Introduction}
Currently, social media and Web 2 tools are integral daily activities in our society because of their popularity and ease of use. Consequently, users of these media generate enormous amounts of data every day. By creating platforms to interact, converse, post ideas, and opinions, social media in its various forms has created a powerful forum for interaction, a way to engage in conversations, and a way to engage in social interaction. In social networks, users often provide useful comments. The different opinions and views that buyers have about a product can reflect how satisfied they are with it and how good it is, acting as a guide for other customers looking to make a similar purchase. Additionally, the results of elections can be predicted by these online comments. The vast number of reviews on one specific topic is difficult to categorize and organize manually. As a result, sentiment analysis emerged as a new field of research due to the need for an automated system for collecting opinions.

Essentially, sentiment analysis is the art of extracting the sentiments expressed in text by applying natural language processing [1-2]. Three levels of sentiment analysis are usually considered in sentiment analysis research: document level, sentence level, and aspect level. It is assumed that the whole text contains opinions only about one subject in the sentiment analysis at the document and sentence levels. This assumption is not logical in many cases [3]. There are different aspects and polarities to opinions in a sentence. We cannot determine a person's true feelings from his or her opinions if the whole sentence includes a positive or negative opinion. In return, Aspect-Based Sentiment Analysis (ABSA), also called Aspect-Level Sentiment Analysis (ALSA), allows us to identify the point of view of the commenter for each of the features of the entity mentioned in the text [3-5].

Research into aspect-based sentiment analysis has been divided into four sub-tasks, which are [6-8]: Aspect Term Extraction (ATE), Aspect Term Polarity (ATP), Aspect Category Detection (ACD), and Aspect Category Polarity (ACP). In contrast to the ATE, which attempts to extract terms that refer to a particular aspect, ATP determines whether terms are positive, negative, or neutral. ACD, on the other hand, intends to assign a subset of these categories to a single review, given a set of predefined categories [4,6,8-15]. The last sub-task of aspect-based sentiment analysis, ACP, focuses on determining the polarity of each category identified by the previous problem [7-8]. Table \ref{tab:1} shows the difference between these four sub-tasks with an example.

\begin{table}[htbp]
\centering
\caption{Comparison between four sub-task of aspect-based sentiment analysis}
\label{tab:1}
\resizebox{\textwidth}{!}{%
\begin{tabular}{|ll|}

\hline
\multicolumn{2}{|l|}{\textbf{Sentence}. Despite the flaws in the script   development, I enjoyed this film, especially the ending, the compelling love   stories, and the message it conveys to the audience.} \\ \hline
\multicolumn{1}{|l|}{\multirow{2}{*}{\textbf{Sub-tasks}}}    & \multirow{2}{*}{\textbf{Output}}                                                                                                                         \\
\multicolumn{1}{|l|}{}                              &                                                                                                                                                 \\ \hline
\multicolumn{1}{|l|}{Aspect Term Extraction}        & script, film, love   stories, message                                                                                                           \\ \hline
\multicolumn{1}{|l|}{Aspect Term Polarity}          & \begin{tabular}[c]{@{}l@{}}\{script: Negative\}, \{film: Positive\},\\    \\ \{love stories: Positive\}, \{message:   Positive\}\end{tabular}   \\ \hline
\multicolumn{1}{|l|}{Aspect Category   Detection}   & Screenplay, Movie,   Story, Content                                                                                                             \\ \hline
\multicolumn{1}{|l|}{Aspect Category   Polarity}    & \begin{tabular}[c]{@{}l@{}}\{Screenplay: Negative\}, \{Movie:   Positive\},\\    \\ \{Story: Positive\}, \{Content:   Positive\}\end{tabular}   \\ \hline
\end{tabular}%
}
\end{table}
In the field of sentiment analysis on reviews, a lot of work has been done. However, due to several challenges, relatively little work has been done in the field of aspect sentiment classification [11]. Since the categories of an aspect are not explicitly mentioned in the text most of the time, the problem of identifying aspect category is a more challenging problem than extracting the aspect term [4,7,11]. Because the term aspect is explicitly stated in the sentence, this is much easier than identifying aspect category [15].
 Most research on aspect-based sentiment analysis has been conducted in a supervised manner. Researchers often extract features and use machine learning algorithms to train the model. These methods are based on the extraction of features manually. However, such methods require feature engineering or extensive linguistic resources. In recent years, deep neural networks, known as deep learning, have shown remarkable performance in various natural language processing tasks, such as text classification, text summarization, question and answer systems, and many more. The principal advantage of these methods is that they can learn text features automatically [16]. Long Short-Term Memory (LSTM) and Convolutional Neural Network (CNN) are two widely used networks in this field.

Most of the previous methods only focus on solving one of these sub-tasks separately. Pipeline methods first identify aspects and then identify polarity. Such methods do not meet practical applications because they can lead to model errors. That is, ACD errors are transmitted to ACP. Due to the importance of aspect-based sentiment analysis, especially its two sub-tasks, ACD and ACP, this article focuses on them. Since most of the previous work deals with solving these two sub-tasks separately, this research presents a new strategy using multi-task deep learning to aspect-based sentiment analysis in Persian comments for the two sub-tasks ACD and ACP, which can solve them simultaneously. Also, since creating a model alone may not provide the best predictions, ensemble learning has been used to enhance the model. The proposed method was evaluated using a data set from the comments provided on Persian movies. Following, after reviewing the relevant works and problem background, we describe in detail the proposed method and the evaluation results.

\section{Background}
\subsection{DEEP LEARNING}
Natural language processing has greatly been influenced by recent advancements in artificial neural networks, particularly deep learning. The use of deep learning has generated results that are superior to those obtained through traditional methods of statistical analysis and machine learning in many areas of natural language processing [17]. In the traditional machine learning method, features were extracted manually and with feature engineering, which requires background knowledge and is also a time-consuming and costly process. Deep learning, also known as feature learning or representation learning, is a subfield of machine learning. Feature detection is performed by a machine based on representation learning, without requiring any human intervention. This critical ability of deep networks, which makes them superior to other traditional machine learning methods, is made possible by many layers in its structure. Deep networks became popular due to their automatic extraction capability and replaced traditional machine learning methods.
\subsubsection{CONVOLUTIONAL NEURAL NETWORKS (CNN)}
\begin{figure}[htbp]
\includegraphics[width=\textwidth]{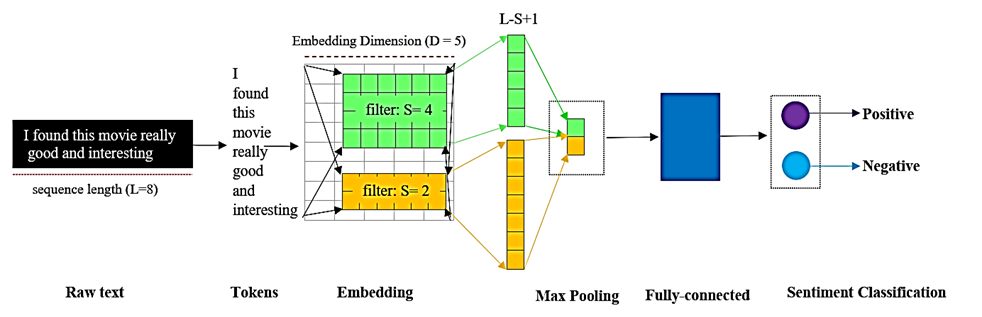}
      \centering
      \caption{The architecture of a CNN model for sentiment classification [20].}
      \label{fig:1}
\end{figure}

Convolutional neural networks (CNNs) are a branch of deep neural networks suitable for processing data with a grid-like topology. A CNN network consists of several layers of convolutional and pooling, followed by one or more fully connected layers [18-19]. CNN's are common in computer vision due to the repetitive patterns in the images. They're also used to process natural language because the text contains repeating patterns as well. Instead of image pixels, the text input in natural language processing is represented as a matrix, with each row representing a word as a vector [18]. These vectors are usually obtained by word embedding or one-hot embedding. Fig. \ref{fig:1}  shows a model of a convolutional neural network architecture for text (sentiment) classification. As the figure shows, the input to this network is a matrix in which the rows represent the vector for each word, and the number of columns represents the vector dimensions of the word. Feature vectors are obtained by applying two filters in sizes 2 and 4 to the input matrix. Once the feature vector is obtained, the max-pooling operator is applied to reduce the size of the features. Finally, the classification is done using a fully connected neural network layer and an activating function.
\subsection{WORD EMBEDDING}
Word embedding [21-22] is a representation of real vectors mapped to words. These numerical vectors are smaller than statistical approaches in natural language processing to convert words into numbers. Also, these numerical vectors, if well trained, can show semantic and syntactic connections between words. Word embedding is the cornerstone of much natural language processing work that uses deep learning. Word embedding can be achieved in two general ways. The first approach is to be learned simultaneously with the main task by the network structure. The second approach is to use pre-trained embedding that has already been trained by specific algorithms.
\subsection{MULTI-TASK LEARNING}
Traditional machine learning methods are optimization for a specific measure for performing a single task. To achieve this goal, a model is trained by fine-tuning the hyperparameters. Although we can get a satisfactory result by training the model, some information that helps to improve the performance of the model is ignored. In other words, we ignore the knowledge gained from training signals related to related tasks. To make effective use of this information, a new approach called multi-task learning was proposed [23-26]. Multi-task learning is done with the aim of joint learning and simultaneously several different and at the same time-related tasks to maximize the efficiency of the model. This is done by sharing information between different tasks. Each task can benefit from other tasks, thus increasing the model's efficiency. In general, whenever you train an optimization problem with more than one loss function, or part of your loss function stems from another task, you are using multi-task learning. [23,26-28]. In matters related to classification, two concepts are easily confused: multi-class classification, which means classification with more than two classes, and multi-label classification, which assigns sets of target tags to each instance. Since both have the prefix "multi," it may be mistaken for both to be multi-task learning. However, this is not true, and when it comes to multi-task learning, we refer to multi-label classification [27].

\subsubsection{MULTI-TASK DEEP LEARNING}
\begin{figure}[htbp]
\includegraphics[width=\textwidth]{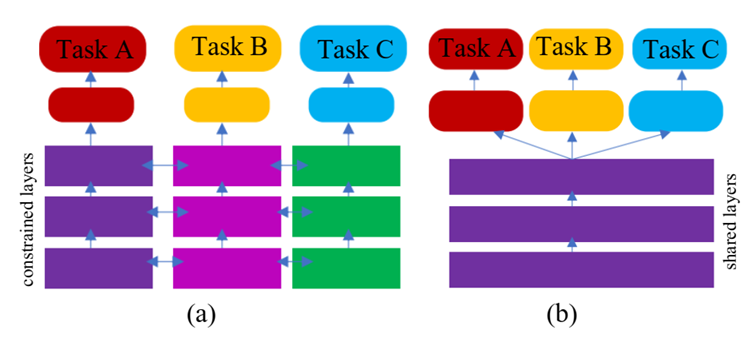}
      \centering
      \caption{Soft parameter sharing (a) and Hard parameter sharing (b) [24].}
      \label{fig:2}
\end{figure}

To perform multi-task learning in deep learning, two general approaches of hard parameter sharing and soft parameter sharing are used. Hard parameter sharing is the most common method of multi-task deep learning [24]. In this method, hidden layers are shared between all tasks, while there are several output layers for each task. In this way, the parameters in the common hidden layer are forced to be generalized to all tasks. This results in a lower overfitting risk for each specific task. In other words, the more tasks are trained simultaneously the model has to find a representation that represents all the tasks and reduces the occurrence of overfitting in the main task. Soft parameter, on the other hand, each task has its model with its parameters. In other words, in soft parameter sharing, each model has its set of weights and biases. In contrast, the distance between these parameters is adjusted in different models to make the parameters similar [23-24,26-27]. Fig. \ref{fig:2} shows the structures of these two methods.
\subsubsection{MULTI-TASK DEEP LEARNING}
\subsection{ENSEMBLE LEARNING}
Ensemble learning is a technique in machine learning that tries to increase the performance of a model compared to single models by combining several models into one model [31][32][33]. The basic principle behind ensemble models is that a group of weak learners come together to help each other form a strong learner. Different approaches to ensemble learning have evolved to lead to better generalizations of learning models. These approaches can be classified into four categories [31]: voting, boosting, bagging, and stacking. Since the voting approach has been used in this research, we will only describe this approach.
\subsubsection{MAJORITY VOTING}
The voting classifier is an ensemble learning approach that is trained with a set of multiple models and predicts class output based on the highest probability of class selection. Each model predicts a result for a sample of input data, which is considered as a vote in favor of the class that the model predicted. When each model predicts an outcome, the final prediction is based on voting for a particular class. The voting classifier supports two types of voting: hard voting and soft voting. In hard voting, known as majority voting, the predicted output class is the majority vote class; That is, the class that was most likely to be predicted by each of the classifiers. In hard voting to prevent equality of votes, it is better to consider the number of classifiers as an odd number. In soft voting, the output class is a prediction based on the average probability of each classifier relative to that class.
\section{RELATED WORK}
Kumar et al. [11] used an association rule technique to identify the aspect categories. To deal with the limitations of these statistical rules, they proposed an approach based on Combined rules, which consists of associative rules and semantic rules. They used word embedding for semantic rules. In the end, they concluded that adding these additional rules and combining them would increase the accuracy of the classification. Ghaderi et al. [9] presented a supervised model called Language-Independent Category Detector (LICD), which detects aspect categories based on text matching without special tools or extracting manual features. To do this, they developed their model with two assumptions. The first assumption is that if a high semantic similarity between a sentence and a set of words representing that category, it should be assigned to a category. And second, a sentence will belong to a category if the sentences are semantically and structurally similar to a sentence related to a category. To apply the assumptions, they used the soft cosine distance for the first assumption and the word mover’s distance for the second assumption. Movahedi et al. [13] presented a deep learning model based on the attention mechanism that can identify different categories by focusing on different parts of a sentence. Xue et al. [12] proposed a model based on the convolutional neural network and the gate mechanism, which performed better than the LSTM. The main idea of this method is to use Tanh-ReLU gate units, which can selectively produce polarity output depending on the aspect and entity. To solve the problems related to supervised approaches that require labeled data, Fu et al. [30] proposed a semi-supervised approach based on the variational autoencoder and the attention mechanism. In addition, to better learn a word, they added the sentiment vector of each word along with it to the input. Their results showed that their models could obtain more accurate sentiments for a given aspect. Ruder et al. [29] proposed a hierarchical model for aspect-based sentiment analysis. They demonstrated that by allowing the model to consider the structure of the review and the sentential context for its predictions, it could perform better than models that rely solely on sentence information and achieve competitive performance with models that use large external sources.

\section{PROPOSED APPROACH }
\begin{figure}[htbp]
\includegraphics[]{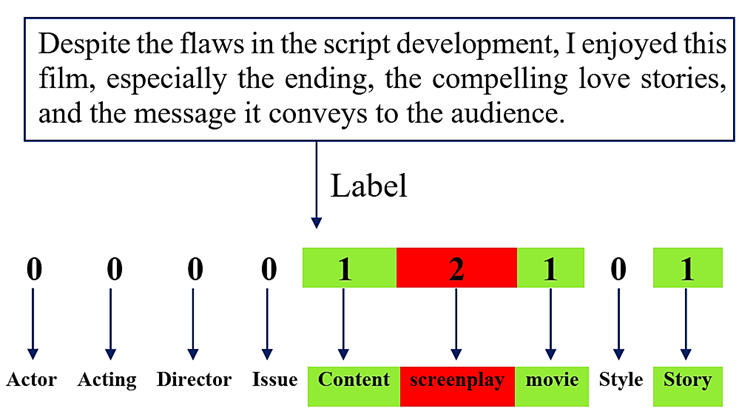}
      \centering
      \caption{An example of labeling in the proposed method.}
      \label{fig:3}
\end{figure}

\begin{figure}[htbp]
\includegraphics[]{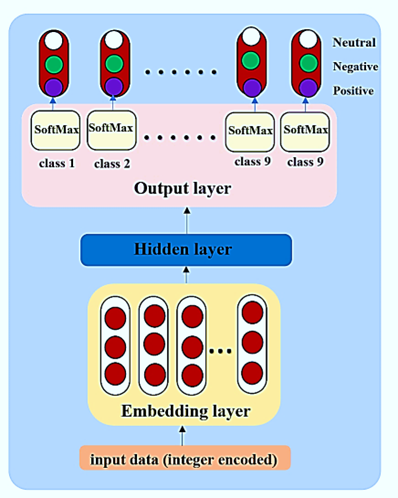}
      \centering
      \caption{The general architecture of the proposed method}
      \label{fig:4}
\end{figure}

\begin{figure}[htbp]
\includegraphics[width=\textwidth]{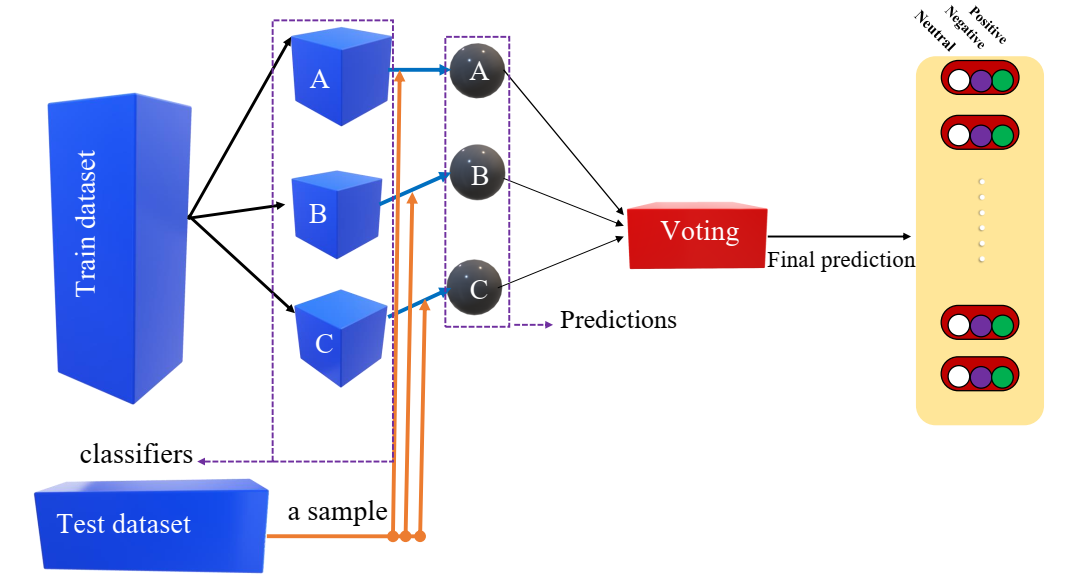}
      \centering
      \caption{majority voting classifier.}
      \label{fig:5}
\end{figure}

The proposed method, by changing the data labeling, or in other words, adding a neutral class gives the model the ability to simultaneously solve the two sub-tasks of aspect category detection and aspect category polarity using deep multi-task learning. To do this, we do not change the aspect category classes and only add a neutral class to the aspect polarity classes. Accordingly, each class of the aspect polarity has three classes: positive, negative, and neutral. Neutral here means that the comment provided is not a member of this class. The reason for adding this neutral class is that it is not possible to judge the final classification by having only two classes of positive and negative polarity. More precisely, since this corresponds to multi-label classification, and it is possible that each instance is a member of one or more of these classes, then we create a third class so that if the instance did not belong to that class, the classifier could detect it. Assuming that a neutral class does not exist, the probability of being positive or negative should be considered for each aspect category. That is, an opinion must have this aspect in the category of positive or negative. It can be seen that this is not a good idea. By adding this neutral class, the model can recognize if a comment is a member of that class or not. Fig. \ref{fig:3} shows an example of how to label using this method for better understanding.
In designing the model for the proposed method, the SoftMax function must be used in the output layer of the model because each category has three different classes of positive, negative, and neutral. Since our number of categories is 9, 9 SoftMax functions with three neurons were used. Fig. \ref{fig:4} shows the general architecture of the proposed method based on hard parameter sharing to solve the two sub-tasks of aspect category detection and aspect category polarity for joint learning.
The proposed method uses an ensemble of multi-task deep convolutional neural networks to improve the sentiment analysis in Persian Reviews. The idea of combining different classifiers starts from the principle that there is no perfect machine learning algorithm. Each model has its limitations and weaknesses, so the goal is to make the best possible decision through a set of algorithms and overcome these limitations. Combining several algorithms is at least more accurate than random guessing because random errors cancel each other out and correct decisions are reinforced. To do this, we have used three CNN models with the same architecture. The only difference between these three CNN models is the size of the word embedding layer. In other words, each of these models learns different textual features, and finally, by combining these features, the final model is created using the majority voting classifier to use all of these features learned by each of the models and the overall error rate of the model. Reduce and thus improve the performance of the educational model. Fig. \ref{fig:5} shows the architecture of the proposed method using majority voting classifier.
The CNN model used in this study consists of several steps: Initially, we first fed the data to the word embedding layer using the Keras library. In the next step, the convolution operator is applied to the word embedding layer, in which 256 filters with a kernel size of 3 were used to form the final features of the input sample. Finally, a fully connected layer was used to classify based on existing classes. 

\section{EXPERIMENT}
\subsection{DATASET}
Every machine learning system needs a suitable data set to learn. Unfortunately, in Persian, there is no comprehensive and unified data set in aspect-based sentiment analysis. Therefore, for this study, a collection of data from the opinions of social media users about movies was collected from cinematicket.org. The number of samples for the train data set is 2000 and for the test data set is 200 samples with nine classes in both positive and negative polarities. Table \ref{tab:2}  shows the distribution of data in different classes.
\begin{table}[htbp]
\centering
\caption{Number of train and test data sets}
\label{tab:2}
\begin{tabular}{|ll|l|l|l|l|l|l|l|l|l|}
\hline
\multicolumn{2}{|l|}{Class/Class}                       & Actor & Acting & Story & Style & movie & screenplay & Content & Issue & Director \\ \hline
\multicolumn{1}{|l|}{\multirow{2}{*}{Positive}} & train & 152   & 200    & 61    & 59    & 907   & 61         & 38      & 56    & 77       \\ \cline{2-11} 
\multicolumn{1}{|l|}{}                          & test  & 11    & 17     & 5     & 16    & 85    & 7          & 7       & 9     & 7        \\ \hline
\multicolumn{1}{|l|}{\multirow{2}{*}{Negative}} & train & 67    & 110    & 72    & 68    & 831   & 111        & 109     & 69    & 62       \\ \cline{2-11} 
\multicolumn{1}{|l|}{}                          & test  & 9     & 7      & 7     & 3     & 61    & 8          & 14      & 6     & 5        \\ \hline
\end{tabular}%
\end{table}
\subsection{EVALUATION METRICS}
Since the topic under discussion is multi-label classification, we must use appropriate measures to evaluate the models. These measures are different from single-label classification measures. In this regard, we have used two measures of jacquard index (multi-label accuracy), and Hamming loss, to evaluate the efficiency of the models.

The Jaccard index, also known as multi-label accuracy, is defined as the number of correctly predicted labels divided by the actual label union. Equation \ref{eq:jaccard} shows how this evaluation measure is calculated.
\begin{equation}
\label{eq:jaccard}
\text{JaccardIndex}(h) = \frac{1}{n} \sum_{i=1}^{n} \frac{|h(x_i) \cap Y_i|}{|h(x_i) \cup Y_i|}
\end{equation}

Hamming loss calculates the symmetric difference between the actual output and the classifier output. In other words, it shows a fraction of the labels that were not correctly predicted for a sample. Equation \ref{eq:hammingloss} shows how this criterion is calculated.
\begin{equation}
\label{eq:hammingloss}
\text{HammingLoss}(h) = \frac{1}{n} \sum_{i=1}^{n} [h(x_i) \oplus Y_i]
\end{equation}

In Equations  \ref{eq:jaccard} to \ref{eq:hammingloss}, Yi represents the actual output for sample i, h(xi) the classifier output, and n represents the number of samples in the data set.
\subsection{Experimental Results}
All experiments this article were performed in the Google Colab environment with the GPU enabled. The Keras library on the TensorFlow platform has been used to build and train the proposed model on dataset. Table \ref{tab:3} compares the models based on the two measures of Jacquard index, and Hamming loss. As can be seen, the proposed model has better results in measures of Jacquard index, and Hamming loss than the models trained as a unit model.
\begin{table}[htbp]
\centering
\caption{Results of experiments with different models on test data sets.}
\label{tab:3}
\begin{tabular}{|l|l|l|l|}
\hline
Model                   & Dimensionality of word & Jaccard Index    & Hamming Loss   \\ \hline
\multirow{3}{*}{CNN}    & 100                    & 77.67\%          & 0.047          \\ \cline{2-4} 
                        & 200                    & 79.28\%          & 0.043          \\ \cline{2-4} 
                        & 300                    & 80.33\%          & 0.041          \\ \hline
\textbf{proposed model} & 100,200,300            & \textbf{\%87.65} & \textbf{0.039} \\ \hline
\end{tabular}%
\end{table}
\subsection{CONCLUSION}
In this article, we used a simple but effective strategy to joint learn the two important sub-task of aspect-based sentiment analysis, namely, ACD and ACP in Persian. In this regard, we used a change in data labeling, or in other words, the addition of a neutral class to enable the multi-task deep learning model to detect category and polarity. Also, in order to increase the efficiency of deep learning models in solving the problem of sentiment analysis in Persian, ensemble learning based on the majority voting approach was used. In this regard, three basic convolutional neural networks were created which are architecturally the same and only the parameter of the dimensions of their embedding layer is different. Experimental results show that the deep learning model based on ensemble learning shows better results compared to the unit models. In other words, combining several models together has been able to reduce each other's random errors and strengthen the right decision.

\section{References}
1.  Nazir A, Rao Y, Wu L, Sun L. “Issues and Challenges of Aspect-based Sentiment Analysis: A Comprehensive Survey”. IEEE Transactions on Affective Computing, (2020).\\
2.  Capuano N, Greco L, Ritrovato P, Vento M. “Sentiment analysis for customer relationship management: an incremental learning approach”. Applied Intelligence, (2020). \\
3.  Zhou J, Huang JX, Chen Q, Hu QV, Wang T, He L. “Deep Learning for Aspect-Level Sentiment Classification: Survey, Vision, and Challenges”. IEEE Access, (2019).\\
4.  Zhu P, Chen Z, Zheng H, Qian T. “Aspect Aware Learning for Aspect Category Sentiment Analysis”. ACM Transactions on Knowledge Discovery from Data, (2019). \\
5.  V H, Al E. “Ensemble Models for Aspect Category Related ABSA Subtasks”. Turkish Journal of Computer and Mathematics Education, (2021).\\
6.  Brychcín T, Konkol M, Steinberger J. UWB: “Machine Learning Approach to Aspect-Based Sentiment Analysis”. Proceedings of the 8th International Workshop on Semantic Evaluation, (2014).\\
7.  Laskari N, Sanampudi S. “Aspect Based Sentiment Analysis Survey” , (2016). \\
8.  Pathak AR, Agarwal B, Pandey M, Rautaray S. “Application of Deep Learning Approaches for Sentiment Analysis”. Algorithms for Intelligent Systems, (2020). \\
9.  Ghadery E, Movahedi S, Jalili Sabet M, Faili H, Shakery A. LICD: “A Language-Independent Approach for Aspect Category Detection”. Lecture Notes in Computer Science, (2019).\\
10.  Dilawar N, Majeed H, Beg MO, Ejaz N, Muhammad K, Mehmood I, et al. “Understanding Citizen Issues through Reviews: A Step towards Data Informed Planning in Smart Cities”. Applied Sciences, (2018). \\
11. Kumar A, Saini M, Sharan A. “Aspect category detection using statistical and semantic association”. Computational Intelligence, (2020).\\
12.  Xue W, Li T. “Aspect Based Sentiment Analysis with Gated Convolutional Networks”. Proceedings of the 56th Annual Meeting of the Association for Computational Linguistics, (2018).\\
13.  Movahedi S, Ghadery E, Faili H, Shakery A. “Aspect Category Detection via Topic-Attention Network” , (2019).\\
14.  babu my, reddy pvp, bindu cs. “aspect category polarity detection using multi class support vector machine with lexicons based features and vector based features”, (2020). \\
15.  Li Y, Yin C, Zhong S. “Sentence Constituent-Aware Aspect-Category Sentiment Analysis with Graph Attention Networks”, (2021).\\
16.  Zhu P, Qian T. “Enhanced Aspect Level Sentiment Classification with Auxiliary Memory”, (2018).\\
17.  Otter DW, Medina JR, Kalita JK. “A Survey of the Usages of Deep Learning in Natural Language Processing”, (2019).\\
18.  Sorin V, Barash Y, Konen E, Klang E. “Deep Learning for Natural Language Processing in Radiology—Fundamentals and a Systematic Review”. Journal of the American College of Radiology, (2020). \\
19.  HARSHA KADAM S. “Text analysis for email multi label classification, MSc Thesis, university of Gothenburg”, (2020).\\
20.  Vazan M, Razmara J. “Jointly Modeling Aspect and Polarity for Aspect-based Sentiment Analysis in Persian Reviews”. 10.13140/RG.2.2.12339.14887/1, (2021).\\
21.  Naili, Marwa, Anja Habacha Chaibi, and Henda Hajjami Ben Ghezala. “Comparative study of word embedding methods in topic segmentation”. Procedia computer science 112 (2017): 340-349. \\
22.  Lombardo, Gianfranco, et al. “Mobility in Unsupervised Word Embeddings for Knowledge Extraction—The Scholars’ Trajectories across Research Topics”. Future Internet 14.1 (2022): 25.\\
23.  Wang S, Wang Q, Gong M. “Multi-Task Learning Based Network Embedding”. Frontiers in Neuroscience, (2020).\\
24.  Ruder S. “An Overview of Multi-Task Learning in Deep Neural Networks”, (2017).\\
25.  Vithayathil Varghese N, Mahmoud QH. “A Survey of Multi-Task Deep Reinforcement Learning”. Electronics, (2020).\\
26.  Ruder S. “Neural Transfer Learning for Natural Language Processing, PhD Thesis, National University of Ireland”, (2019).\\
27.  Xin W. “Multi-Task Deep Learning for Affective Content Detection from Text, MSc Thesis, University of Ottawa”, (2020).\\
28.  Gong T, Lee T, Stephenson C, Renduchintala V, Padhy S, Ndirango A, et al. “A Comparison of Loss Weighting Strategies for Multi task Learning in Deep Neural Networks”. IEEE Access, (2019). \\
29.  Ruder S, Ghaffari P, Breslin JG. “A Hierarchical Model of Reviews for Aspect-based Sentiment Analysis”, (2016).\\
30.  Fu X, Wei Y, Xu F, Wang T, Lu Y, Li J, et al. “Semi-supervised Aspect-level Sentiment Classification Model based on Variational Autoencoder. Knowledge-Based Systems”, (2019).\\
31.  Ganaie M.A, Hu M, Tanveer M, Suganthan P.N. “Ensemble deep learning: A review”. arXiv:2104.02395 [cs], (2021).\\
32.  Mohammadi A, Shaverizade A. “Ensemble deep learning for aspect-based sentiment analysis”, International Journal of Nonlinear Analysis and Applications, pp.29-38, (2021). \\
33.  Sagi O, Rokach L. “Ensemble learning: A survey”. Wiley Interdisciplinary Reviews: Data Mining and Knowledge Discovery, (2018).\\

\end{document}